\documentclass[10pt,a4paper,onecolumn]{article}
\usepackage{marginnote}
\usepackage{graphicx}
\usepackage{xcolor}
\usepackage{authblk,etoolbox}
\usepackage{titlesec}
\usepackage{calc}
\usepackage{tikz}
\usepackage{hyperref}
\hypersetup{colorlinks,breaklinks=true,
            urlcolor=[rgb]{0.0, 0.5, 1.0},
            linkcolor=[rgb]{0.0, 0.5, 1.0}}
\usepackage{caption}
\usepackage{tcolorbox}
\usepackage{amssymb,amsmath}
\usepackage{ifxetex,ifluatex}
\usepackage{seqsplit}
\usepackage{xstring}

\usepackage{float}
\let\origfigure\figure
\let\endorigfigure\endfigure
\renewenvironment{figure}[1][2] {
    \expandafter\origfigure\expandafter[H]
} {
    \endorigfigure
}

\usepackage{fixltx2e} % provides \textsubscript
\usepackage[
  backend=biber,
]{biblatex}
\bibliography{paper.bib}

\let\textttOrig=\texttt
\def\texttt#1{\expandafter\textttOrig{\seqsplit{#1}}}
\renewcommand{\seqinsert}{\ifmmode
  \allowbreak
  \else\penalty6000\hspace{0pt plus 0.02em}\fi}

\makeatletter
\let\href@Orig=\href
\def\href@Urllike#1#2{\href@Orig{#1}{\begingroup
    \def\Url@String{#2}\Url@FormatString
    \endgroup}}
\def\href@Notdoi#1#2{\def\tempa{#1}\def\tempb{#2}%
  \ifx\tempa\tempb\relax\href@Urllike{#1}{#2}\else
  \href@Orig{#1}{#2}\fi}
\def\href#1#2{%
  \IfBeginWith{#1}{https://doi.org}%
  {\href@Urllike{#1}{#2}}{\href@Notdoi{#1}{#2}}}
\makeatother

\newlength{\cslhangindent}
\newlength{\csllabelwidth}
\newenvironment{CSLReferences}[3] % #1 hanging-ident, #2 entry spacing
 {% don't indent paragraphs
  \setlength{\parindent}{0pt}
  \ifodd #1 \everypar{\setlength{\hangindent}{\cslhangindent}}\ignorespaces\fi
  \ifnum #2 > 0
  \setlength{\parskip}{#2\baselineskip}
  \fi
 }%
 {}
\usepackage{calc}

\usepackage[top=3.5cm, bottom=3cm, right=1.5cm, left=1.0cm,
            headheight=2.2cm, reversemp, includemp, marginparwidth=4.5cm]{geometry}

\titleformat{\section}
  {\normalfont\sffamily\Large\bfseries}
  {}{0pt}{}
\titleformat{\subsection}
  {\normalfont\sffamily\large\bfseries}
  {}{0pt}{}
\titleformat{\subsubsection}
  {\normalfont\sffamily\bfseries}
  {}{0pt}{}
\titleformat*{\paragraph}
  {\sffamily\normalsize}

\usepackage{fancyhdr}
\makeatletter
\let\ps@plain\ps@fancy
\fancyheadoffset[L]{4.5cm}
\fancyfootoffset[L]{4.5cm}

\definecolor{linky}{rgb}{0.0, 0.5, 1.0}

\newtcolorbox{repobox}
   {colback=red, colframe=red!75!black,
     boxrule=0.5pt, arc=2pt, left=6pt, right=6pt, top=3pt, bottom=3pt}

\newcommand{\ExternalLink}{%
   \tikz[x=1.2ex, y=1.2ex, baseline=-0.05ex]{%
       \begin{scope}[x=1ex, y=1ex]
           \clip (-0.1,-0.1)
               --++ (-0, 1.2)
               --++ (0.6, 0)
               --++ (0, -0.6)
               --++ (0.6, 0)
               --++ (0, -1);
           \path[draw,
               line width = 0.5,
               rounded corners=0.5]
               (0,0) rectangle (1,1);
       \end{scope}
       \path[draw, line width = 0.5] (0.5, 0.5)
           -- (1, 1);
       \path[draw, line width = 0.5] (0.6, 1)
           -- (1, 1) -- (1, 0.6);
       }
   }

\patchcmd{\@maketitle}{center}{flushleft}{}{}
\patchcmd{\@maketitle}{center}{flushleft}{}{}
\patchcmd{\@maketitle}{\LARGE}{\LARGE\sffamily}{}{}
\def\maketitle{{%
  
  \AB@maketitle}}
\makeatletter
\renewcommand\AB@affilsepx{ \protect\Affilfont}
\renewcommand\AB@affilnote[1]{{\bfseries #1}\hspace{3pt}}
\renewcommand{\affil}[2][]%
   {\newaffiltrue\let\AB@blk@and\AB@pand
      \if\relax#1\relax\def\AB@note{\AB@thenote}\else\def\AB@note{#1}%
        \setcounter{Maxaffil}{0}\fi
        \begingroup
        \let\href=\href@Orig
        \let\texttt=\textttOrig
        \let\protect\@unexpandable@protect
        \def\thanks{\protect\thanks}\def\footnote{\protect\footnote}%
        \@temptokena=\expandafter{\AB@authors}%
        {\def\\{\protect\\\protect\Affilfont}\xdef\AB@temp{#2}}%
         \xdef\AB@authors{\the\@temptokena\AB@las\AB@au@str
         \protect\\[\affilsep]\protect\Affilfont\AB@temp}%
         \gdef\AB@las{}\gdef\AB@au@str{}%
        {\def\\{, \ignorespaces}\xdef\AB@temp{#2}}%
        \@temptokena=\expandafter{\AB@affillist}%
        \xdef\AB@affillist{\the\@temptokena \AB@affilsep
          \AB@affilnote{\AB@note}\protect\Affilfont\AB@temp}%
      \endgroup
       \let\AB@affilsep\AB@affilsepx
}
\makeatother

\renewcommand\Affilfont{\sffamily\small\mdseries}
\ifnum 0\ifxetex 1\fi\ifluatex 1\fi=0 % if pdftex
  \usepackage[T1]{fontenc}
  \usepackage[utf8]{inputenc}

\else % if luatex or xelatex
  \ifxetex
    \usepackage{mathspec}
    \usepackage{fontspec}

  \else
    \usepackage{fontspec}
  \fi
  \defaultfontfeatures{Ligatures=TeX,Scale=MatchLowercase}

\fi
\IfFileExists{upquote.sty}{\usepackage{upquote}}{}
\IfFileExists{microtype.sty}{%
\usepackage{microtype}
\UseMicrotypeSet[protrusion]{basicmath} % disable protrusion for tt fonts
}{}

\usepackage{hyperref}
\hypersetup{unicode=true,
            pdftitle={FastGeodis: Fast Generalised Geodesic Distance Transform},
            pdfborder={0 0 0},
            breaklinks=true}
\let\addcontentslineOrig=\addcontentsline
\def\addcontentsline#1#2#3{\bgroup
  \let\texttt=\textttOrig\addcontentslineOrig{#1}{#2}{#3}\egroup}
\let\markbothOrig\markboth
\def\markboth#1#2{\bgroup
  \let\texttt=\textttOrig\markbothOrig{#1}{#2}\egroup}
\let\markrightOrig\markright
\def\markright#1{\bgroup
  \let\texttt=\textttOrig\markrightOrig{#1}\egroup}

\usepackage{graphicx,grffile}
\makeatletter
\def\maxwidth{\ifdim\Gin@nat@width>\linewidth\linewidth\else\Gin@nat@width\fi}
\def\maxheight{\ifdim\Gin@nat@height>\textheight\textheight\else\Gin@nat@height\fi}
\makeatother
\setkeys{Gin}{width=\maxwidth,height=\maxheight,keepaspectratio}
\IfFileExists{parskip.sty}{%
\usepackage{parskip}
}{% else
\setlength{\parindent}{0pt}
\setlength{\parskip}{6pt plus 2pt minus 1pt}
}
\ifx\paragraph\undefined\else
\let\oldparagraph\paragraph
\renewcommand{\paragraph}[1]{\oldparagraph{#1}\mbox{}}
\fi
\ifx\subparagraph\undefined\else
\let\oldsubparagraph\subparagraph
\renewcommand{\subparagraph}[1]{\oldsubparagraph{#1}\mbox{}}
\fi

\title{FastGeodis: Fast Generalised Geodesic Distance Transform}

        \author[1]{Muhammad Asad}
          \author[1]{Reuben Dorent}
          \author[1]{Tom Vercauteren}
    
      \affil[1]{School of Biomedical Engineering \& Imaging Sciences,
King's College London, UK}
  \date{\vspace{-7ex}}

\begin{document}
\maketitle

\marginpar{

  \begin{flushleft}
  %\hrule
  \sffamily\small

  {\bfseries DOI:} \href{https://doi.org/10.21105/joss.04532}{\color{linky}{10.21105/joss.04532}}

  \vspace{2mm}

  {\bfseries Software}
  \begin{itemize}
    \setlength\itemsep{0em}
    \item \href{https://github.com/openjournals/joss-reviews/issues/4532}{\color{linky}{Review}} \ExternalLink
    \item \href{https://github.com/masadcv/FastGeodis}{\color{linky}{Repository}} \ExternalLink
    \item \href{https://doi.org/10.5281/zenodo.7349069}{\color{linky}{Archive}} \ExternalLink
  \end{itemize}

  \vspace{2mm}

  \par\noindent\hrulefill\par

  \vspace{2mm}

  {\bfseries Editor:} \href{https://vissarion.github.io/}{Vissarion Fisikopoulos} \ExternalLink \\
  \vspace{1mm}
    {\bfseries Reviewers:}
  \begin{itemize}
  \setlength\itemsep{0em}
    \item \href{https://github.com/jacobmerson}{@jacobmerson}
    \item \href{https://github.com/dataplayer12}{@dataplayer12}
    \end{itemize}
    \vspace{2mm}

  {\bfseries Submitted:} 27 June 2022\\
  {\bfseries Published:} 23 November 2022

  \vspace{2mm}
  {\bfseries License}\\
  Authors of papers retain copyright and release the work under a Creative Commons Attribution 4.0 International License (\href{http://creativecommons.org/licenses/by/4.0/}{\color{linky}{CC BY 4.0}}).

  \end{flushleft}
}

\hypertarget{summary}{%
\section{Summary}\label{summary}}

Geodesic and Euclidean distance transforms have been widely used in a
number of applications where distance from a set of reference points is
computed. Methods from recent years have shown effectiveness in applying
the Geodesic distance transform to interactively annotate 3D medical
imaging data (Criminisi et al., 2008; Wang et al., 2018). The Geodesic
distance transform enables providing segmentation labels, i.e.,
voxel-wise labels, for different objects of interests. Despite existing
methods for efficient computation of the Geodesic distance transform on
GPU and CPU devices (Criminisi et al., 2008, 2009; Toivanen, 1996; Weber
et al., 2008), an open-source implementation of such methods on the GPU
does not exist. On the contrary, efficient methods for the computation
of the Euclidean distance transform (Felzenszwalb \& Huttenlocher, 2012)
have open-source implementations (Abadi et al., 2015; Seung-Lab, 2018).
Existing libraries, e.g., Wang (2020), provide C++ implementations of
the Geodesic distance transform; however, they lack efficient
utilisation of the underlying hardware and hence result in significant
computation time, especially when applying them on 3D medical imaging
volumes.

The \texttt{FastGeodis} package provides an efficient implementation for
computing Geodesic and Euclidean distance transforms (or a mixture of
both), targeting efficient utilisation of CPU and GPU hardware. In
particular, it implements the paralellisable raster scan method from
Criminisi et al. (2009), where elements in a row (2D) or plane (3D) can
be computed with parallel threads. This package is able to handle 2D as
well as 3D data, where it achieves up to a 20x speedup on a CPU and up
to a 74x speedup on a GPU as compared to an existing open-source library
(Wang, 2020) that uses a non-parallelisable single-thread CPU
implementation. The performance speedups reported here were evaluated
using 3D volume data on an Nvidia GeForce Titan X (12 GB) with a 6-Core
Intel Xeon E5-1650 CPU. Further in-depth comparison of performance
improvements is discussed in the \texttt{FastGeodis}
\href{https://fastgeodis.readthedocs.io/}{documentation}.

\hypertarget{statement-of-need}{%
\section{Statement of need}\label{statement-of-need}}

Despite existing open-source implementation of distance transforms
(Abadi et al., 2015; Seung-Lab, 2018; Wang, 2020), open-source
implementations of efficient Geodesic distance transform algorithms
(Criminisi et al., 2009; Weber et al., 2008) on CPUs and GPUs do not
exist. However, efficient CPU (Seung-Lab, 2018) and GPU (Abadi et al.,
2015) implementations exist for Euclidean distance transform. To the
best of our knowledge, \texttt{FastGeodis} is the first open-source
implementation of efficient the Geodesic distance transform (Criminisi
et al., 2009), achieving up to a 20x speedup on a CPU and up to a 74x
speedup on a GPU as compared to existing open-source libraries (Wang,
2020). It also provides an efficient implementation of the Euclidean
distance transform. In addition, it is the first open-source
implementation of generalised Geodesic distance transform and Geodesic
Symmetric Filtering (GSF) as proposed in Criminisi et al. (2008). Apart
from a method from Criminisi et al. (2009), Weber et al. (2008) present
a further optimised approach for computing Geodesic distance transforms
on GPUs. However, this method is protected by multiple patents
(Bronstein et al., 2013, 2015, 2016) and hence is not suitable for
open-source implementation in the \textbf{FastGeodis} package.

The ability to efficiently compute Geodesic and Euclidean distance
transforms can significantly enhance distance transform applications,
especially for training deep learning models that utilise distance
transforms (Wang et al., 2018). It will improve prototyping,
experimentation, and deployment of such methods, where efficient
computation of distance transforms has been a limiting factor. In 3D
medical imaging problems, efficient computation of distance transforms
will lead to significant speed-ups, enabling online learning
applications for better processing/labelling/inference from volumetric
datasets (Asad et al., 2022). In addition, \texttt{FastGeodis} provides
an efficient implementation for both CPUs and GPUs and hence will enable
efficient use of a wide range of hardware devices.

\hypertarget{implementation}{%
\section{Implementation}\label{implementation}}

\begin{figure}
\centering
\includegraphics[width=0.8\textwidth,height=\textheight]{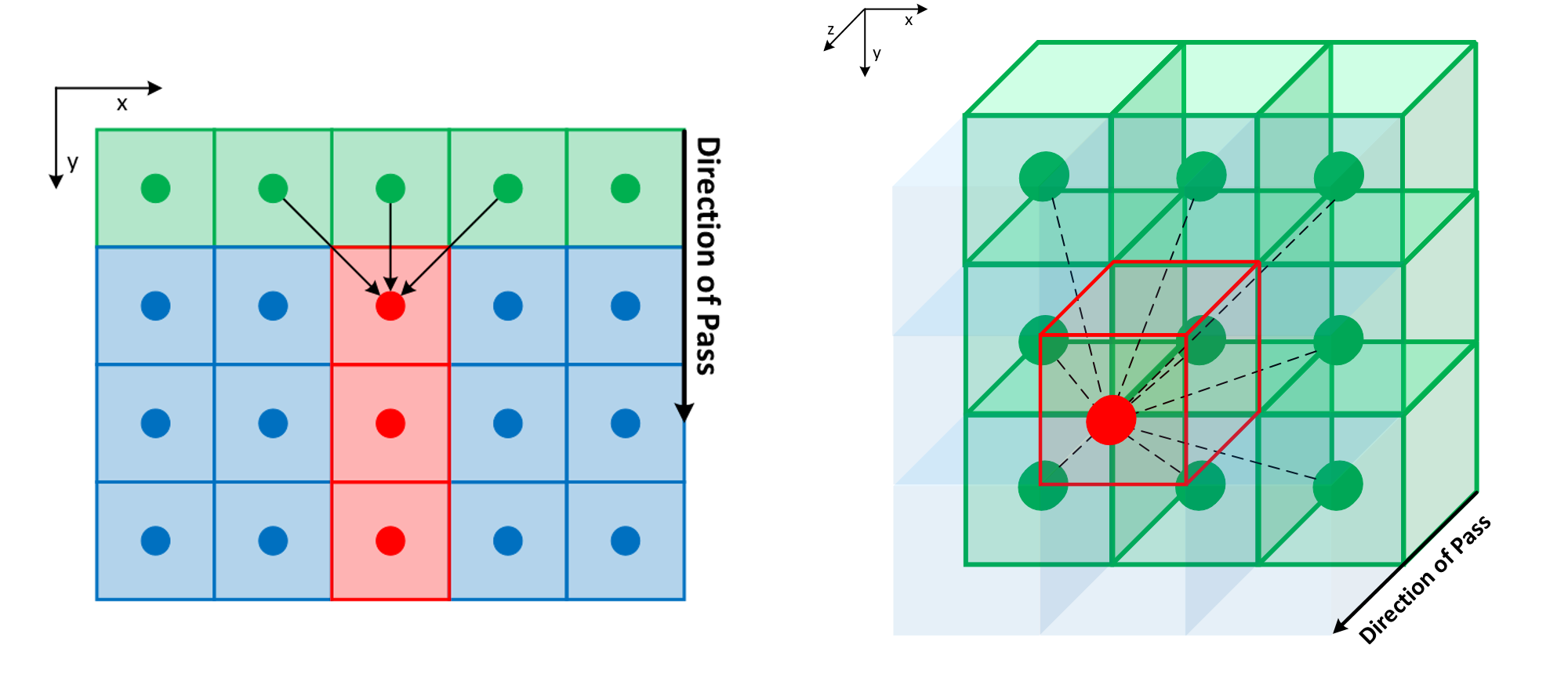}
\caption{Raster scan data propagation passes in
FastGeodis.\label{fig:hwpasses}}
\end{figure}

\texttt{FastGeodis} implements an efficient distance transform algorithm
from Criminisi et al. (2009), which provides parallelisable raster scans
to compute distance transform. The implementation consists of data
propagation passes that are parallelised using threads for elements
across a line (2D) or plane (3D). \autoref{fig:hwpasses} shows these
data propagation passes, where each pass consists of computing distance
values for the next row (2D) or plane (3D) by utilising parallel threads
and data from the previous row/plane, hence resulting in propagating
distance values along the direction of the pass. For 2D data, four
distance propagation passes are required, top-bottom, bottom-top,
left-right and right-left, whereas for 3D data six passes are required,
front-back, back-front, top-bottom, bottom-top, left-right and
right-left. The algorithm can be applied to efficiently compute both
Geodesic and Euclidean distance transforms. In addition to this,
\texttt{FastGeodis} also provides the non-parallelisable raster scan
based distance transform method from Toivanen (1996), which is
implemented using a single CPU thread and used for comparison.

The \texttt{FastGeodis} package is implemented using \texttt{PyTorch}
(Paszke et al., 2019), utilising OpenMP for CPU- and CUDA for
GPU-parallelisation of the algorithm. It is accessible as a Python
package that can be installed across different operating systems and
devices. Comprehensive documentation and a range of examples are
provided for understanding the usage of the package on 2D and 3D data
using CPUs or GPUs. Two- and three-dimensional examples are provided for
Geodesic, Euclidean, and Signed Geodesic distance transforms as well as
for computing Geodesic Symmetric Filtering (GSF), the essential first
step in implementing the interactive segmentation method described in
Criminisi et al. (2008). A further in-depth overview of the implemented
algorithm, along with evaluation on common 2D/3D data input sizes, is
provided in the \texttt{FastGeodis}
\href{https://fastgeodis.readthedocs.io/}{documentation}.

\hypertarget{acknowledgements}{%
\section{Acknowledgements}\label{acknowledgements}}

This research was supported by the European Union's Horizon 2020
research and innovation programme under grant agreement No 101016131.

\hypertarget{references}{%
\section*{References}\label{references}}
\addcontentsline{toc}{section}{References}

\hypertarget{refs}{}
\begin{CSLReferences}{1}{0}
\leavevmode\hypertarget{ref-tensorflow2015-whitepaper}{}%
Abadi, M., Agarwal, A., Barham, P., Brevdo, E., Chen, Z., Citro, C.,
Corrado, G. S., Davis, A., Dean, J., Devin, M., Ghemawat, S.,
Goodfellow, I., Harp, A., Irving, G., Isard, M., Jia, Y., Jozefowicz,
R., Kaiser, L., Kudlur, M., \ldots{} Zheng, X. (2015).
\emph{{TensorFlow}: Large-scale machine learning on heterogeneous
systems}. \url{https://www.tensorflow.org/}

\leavevmode\hypertarget{ref-asad2022econet}{}%
Asad, M., Fidon, L., \& Vercauteren, T. (2022). ECONet: Efficient
convolutional online likelihood network for scribble-based interactive
segmentation. \emph{arXiv Preprint arXiv:2201.04584}.

\leavevmode\hypertarget{ref-bronstein2013parallel}{}%
Bronstein, A., Bronstein, M., Devir, Y., Weber, O., \& Kimmel, R.
(2013). \emph{Parallel approximation of distance maps (US patent
8,373,716)}. Google Patents.

\leavevmode\hypertarget{ref-bronstein2015parallel}{}%
Bronstein, A., Bronstein, M., Kimmel, R., Devir, Y., \& Weber, O.
(2015). \emph{Parallel approximation of distance maps (US patent
8,982,142)}. Google Patents.

\leavevmode\hypertarget{ref-bronstein2016parallel}{}%
Bronstein, A., Bronstein, M., Kimmel, R., Devir, Y., \& Weber, O.
(2016). \emph{Parallel approximation of distance maps (US patent
9,489,708)}. Google Patents.

\leavevmode\hypertarget{ref-criminisi2008geos}{}%
Criminisi, A., Sharp, T., \& Blake, A. (2008). Geos: Geodesic image
segmentation. \emph{European Conference on Computer Vision}, 99--112.
\url{https://doi.org/10.1007/978-3-540-88682-2_9}

\leavevmode\hypertarget{ref-criminisiinteractive}{}%
Criminisi, A., Sharp, T., \& Siddiqui, K. (2009). Interactive {G}eodesic
segmentation of n-dimensional medical images on the graphics processor.
\emph{Radiological Society of North America (RSNA)}.
\url{https://www.microsoft.com/en-us/research/publication/interactive-geodesic-segmentation-of-n-dimensional-medical-images-on-the-graphics-processor/}

\leavevmode\hypertarget{ref-felzenszwalb2012distance}{}%
Felzenszwalb, P. F., \& Huttenlocher, D. P. (2012). Distance transforms
of sampled functions. \emph{Theory of Computing}, \emph{8}(1), 415--428.

\leavevmode\hypertarget{ref-NEURIPS2019_9015}{}%
Paszke, A., Gross, S., Massa, F., Lerer, A., Bradbury, J., Chanan, G.,
Killeen, T., Lin, Z., Gimelshein, N., Antiga, L., Desmaison, A., Kopf,
A., Yang, E., DeVito, Z., Raison, M., Tejani, A., Chilamkurthy, S.,
Steiner, B., Fang, L., \ldots{} Chintala, S. (2019). PyTorch: An
imperative style, high-performance deep learning library. In H. Wallach,
H. Larochelle, A. Beygelzimer, F. dAlché-Buc, E. Fox, \& R. Garnett
(Eds.), \emph{Advances in neural information processing systems 32} (pp.
8024--8035). Curran Associates, Inc.
\url{http://papers.neurips.cc/paper/9015-pytorch-an-imperative-style-high-performance-deep-learning-library.pdf}

\leavevmode\hypertarget{ref-eucildeantdimpl}{}%
Seung-Lab. (2018). Multi-label anisotropic 3D {E}uclidean distance
transform (MLAEDT-3D). In \emph{GitHub repository}. GitHub.
\url{https://github.com/seung-lab/euclidean-distance-transform-3d}

\leavevmode\hypertarget{ref-toivanen1996new}{}%
Toivanen, P. J. (1996). New geodosic distance transforms for gray-scale
images. \emph{Pattern Recognition Letters}, \emph{17}(5), 437--450.
\url{https://doi.org/10.1016/0167-8655(96)00010-4}

\leavevmode\hypertarget{ref-geodistk}{}%
Wang, G. (2020). GeodisTK: Geodesic distance transform toolkit for 2D
and 3D images. In \emph{GitHub repository}. GitHub.
\url{https://github.com/taigw/GeodisTK}

\leavevmode\hypertarget{ref-wang2018deepigeos}{}%
Wang, G., Zuluaga, M. A., Li, W., Pratt, R., Patel, P. A., Aertsen, M.,
Doel, T., David, A. L., Deprest, J., Ourselin, S., \& others. (2018).
DeepIGeoS: A deep interactive geodesic framework for medical image
segmentation. \emph{IEEE Transactions on Pattern Analysis and Machine
Intelligence}, \emph{41}(7), 1559--1572.

\leavevmode\hypertarget{ref-weber2008parallel}{}%
Weber, O., Devir, Y. S., Bronstein, A. M., Bronstein, M. M., \& Kimmel,
R. (2008). Parallel algorithms for approximation of distance maps on
parametric surfaces. \emph{ACM Transactions on Graphics (TOG)},
\emph{27}(4), 1--16. \url{https://doi.org/10.1145/1409625.1409626}

\end{CSLReferences}

\end{document}